# Data Clustering using a Hybrid of Fuzzy C-Means and Quantum-behaved Particle Swarm Optimization

Saptarshi Sengupta
Department of EECS
Vanderbilt University
Nashville, TN, USA
saptarshi.sengupta@vanderbilt.edu

Sanchita Basak
Department of EECS
Vanderbilt University
Nashville, TN, USA
sanchita.basak@vanderbilt.edu

Richard Alan Peters II
Department of EECS
Vanderbilt University
Nashville, TN, USA
alan.peters@vanderbilt.edu

*Abstract* – **Fuzzy clustering has become a widely used data mining technique and plays an important role in grouping, traversing and selectively using data for user specified applications. The deterministic Fuzzy C-Means (FCM) algorithm may result in suboptimal solutions when applied to multidimensional data in real-world, time-constrained problems. In this paper the Quantum-behaved Particle Swarm Optimization (QPSO) with a fully connected topology is coupled with the Fuzzy C-Means Clustering algorithm and is tested on a suite of datasets from the UCI Machine Learning Repository. The global search ability of the QPSO algorithm helps in avoiding stagnation in local optima while the soft clustering approach of FCM helps to partition data based on membership probabilities. Clustering performance indices such as F-Measure, Accuracy, Quantization Error, Intercluster and Intracluster distances are reported for competitive techniques such as PSO K-Means, QPSO K-Means and QPSO FCM over all datasets considered. Experimental results indicate that QPSO FCM provides comparable and in most cases superior results when compared to the others.**

## I. INTRODUCTION

Clustering is the process of grouping sets of objects such that objects in one group are more similar to each other than to those in another group. Data clustering is widely used for statistical analyses in machine learning, pattern recognition, image analysis and the information sciences making it a common exploratory data mining technique [1-2]. The K-Means algorithm is one of the widely used partitioned data clustering techniques, however its solution quality is sensitive to the initial choice of cluster centres and it is susceptible to getting trapped in local optima [1]. K-Means is NP-hard, thus approximation algorithms have been used to obtain close to exact solutions [3]. Fuzzy C-Means (FCM) [4] algorithm is an unsupervised soft clustering approach which uses a membership function to assign an object to multiple clusters but suffers from the same issue of stagnation using iterative gradient descent as in hard K-Means. This has led to several attempts to intelligently traverse the search space and minimize the underlying cost, often at the expense of increased time complexity. In the past two decades, powered by increased computational capabilities and the advent of nature-inspired algorithmic models of collective intelligence and emergence, many studies have led to the application of guided random search algorithms in cost optimization of partitioned and soft clustering. Several metaheuristics mimicking information exchange in social colonies of bird and insect species are well known for their robust performances on ill-structured global optimization problems, irrespective of the continuity or gradient of the cost function. This paper makes a comparative analysis of the performance of one such algorithm: the Quantum-behaved Particle Swarm Optimization (QPSO) [16], from both a hard, partitioned (QPSO K-Means) as well as a soft, fuzzy clustering (FCM QPSO) point of view. The literature suggests prior work on integrating Particle Swarm Optimization (PSO) [8] into the deterministic K-Means framework has led to improved clustering accuracy across many datasets. This is evidenced by the works of Izakian et al. [5], Emami et al. [6] and Yang et al. [7], among others. In [5] the authors integrated FCM with a fuzzy PSO and noted the efficiency and improvement in solution quality whereas the authors of [6] hybridized FCM with PSO on one hand and an Imperialist Competitive Algorithm (ICA) [24] on the other to come to the conclusion that ICAPSO suited the clustering jobs under consideration better than the competitor methods tested. The work of Yang et al. in [7] used as metric the harmonic average of distances between individual data points and cluster centres summed over all points. The proposed PSO K-Harmonic Means (PSOKHM) in [7] was found to outperform K-Harmonic Means (KHM) and PSO in that it not only reduced convergence time of PSO but also helped KHM escape local minima. In this work, a detailed report of performance indices for some popular datasets from the UCI Machine Learning Repository [20] using FCM QPSO is made against QPSO K-Means, PSO K-Means and traditional K-Means. Subsequent sections of the paper are structured as follows: Section II elaborates on the FCM algorithm, Section III introduces the variants of PSO used and Section IV describes the FCM QPSO approach. Section V details the experimental setup while Section VI reports and analyzes the results obtained. Finally, Section VII makes concluding remarks.

## II. FUZZY C-MEANS ALGORITHM (FCM)

The Fuzzy C-Means (FCM) algorithm aims to partition $N$ objects into $C$ clusters. Essentially, this reduces to grouping the object set $D = \{D_1, D_2, D_3 \ldots \ldots D_N\}$ into $C$ clusters ($1 < C < N$) with $\Omega = \{\Omega_1, \Omega_2, \Omega_3, \ldots \Omega_C\}$ being the cluster centres. Each data point belongs to a cluster with randomly initialized centroids, according to a membership function $\mu_{ij}$ defined as:

$$\mu_{ij} = \frac{1}{\sum_{r=1}^{C} (\frac{d_{ij}}{d_{rj}})^{\frac{2}{m-1}}} \quad (1)$$

$d_{ij} = \| x_i - y_j \|$ is the distance between *i-th* centre and *j-th* data point, $d_{rj} = \| x_r - y_j \|$ is that between *r-th* centre and *j-th* data point and $m \in [1, \infty)$ is a fuzzifier. FCM employs an iterative gradient descent to compute centroids, which are updated as:

$$x_i = \frac{\sum_{j=1}^{N} \mu_{ij}^m y_j}{\sum_{j=1}^{N} \mu_{ij}^m} \quad (2)$$

The objective function minimized by FCM can be formulated as the sum of membership weighted Euclidean distances:

$$\varphi = \sum_{i=1}^{C} \sum_{j=1}^{N} \mu_{ij}^m \left( \| x_i - y_j \| \right)^2 \quad (3)$$

By recursively calculating eqs. (1) and (2), FCM can be terminated once a preset convergence criteria is met. Like many algorithms which employ gradient descent, FCM can fall prey to local optima in a multidimensional fitness landscape. To avoid this, a stochastic optimization approach can be used.

## III. VARIANTS OF PARTICLE SWARM OPTIMIZERS USED

### A. Particle Swarm Optimization (PSO)

PSO proposed by Eberhart and Kennedy [8] is a stochastic optimization strategy that makes no assumptions about the gradient of the objective function. It has been able to effectively produce promising results in many engineering problems where deterministic algorithms fail. Although PSO is widely considered a universal optimizer there exist numerous issues with the standard PSO [8], most notably a poor local search ability (Angeline et. al) [9]. This has led to several subsequent studies on improvements of the same [10-13]. The particles in PSO update their position through a personal best position - *pbest* and a global best - *gbest*. After each iteration their velocity and position are updated as:

$$v_{ij}(t+1) = \omega v_{ij}(t) + C_1 r_1(t) \left( p_{ij}(t) - x_{ij}(t) \right)$$
$$+ C_2 r_2(t) \left( p_{gj}(t) - x_{ij}(t) \right) \quad (4)$$

$$x_{ij}(t+1) = x_{ij}(t) + v_{ij}(t+1) \quad (5)$$

$C_1$ and $C_2$ are social and cognitive acceleration constants, $r_1$ and $r_2$ are i.i.d. random numbers between 0 and 1, $x_{ij}$, $v_{ij}$ represent the position and velocity of $i_{th}$ particle in $j_{th}$ dimension whereas $p_{ij}(t)$ and $p_{gj}(t)$ are the *pbest* and *gbest* positions. In term 1 in the RHS of eq. (4), $\omega$ represents the inertia of the *i*-th particle and terms 2 and 3 introduce guided perturbations towards basins of attraction in the direction of movement of the particle. The personal best (*pbest*) update follows a greedy update scheme considering a cost minimization goal, as discussed in the following equation.

$$f(x_i(t+1)) < f(p_i(t)) \Rightarrow p_i(t+1) = x_i(t+1)$$

else $p_i(t+1) = p_i(t)$ (6)

Here, *f* is the cost and $p_i$ is the personal best of a particle. The global best ($p_g$) is the minimum cost bearing element of the historical set of personal bests $p_i$ of a particular particle. A major limitation of the standard PSO is its inability to guarantee convergence to an optimum as was shown by Van den Bergh [14] based on the criterion established in [15].

### B. Quantum-behaved PSO (QPSO)

Sun et al. proposed a delta potential well model for PSO, leading to a variant known as Quantum-behaved Particle Swarm Optimization (QPSO) [16]. A detailed analysis of the derivation of particle trajectories in QPSO may be found in [16-19]. The state update equations of a particle in a fully connected QPSO topology is described by the following equations:

$$mbest_j = \frac{1}{N} \sum_{i=1}^{N} p_{ij} \quad (7)$$

$$\Phi_{ij} = \theta p_{ij} + (1 - \theta) p_{gj} \quad (8)$$

$$x_{ij} = \Phi_{ij} + \beta \left| mbest_j - x_{ij}(t) \right| \ln(1/q) \quad \forall \, k \geq 0.5$$
$$= \Phi_{ij} - \beta \left| mbest_j - x_{ij}(t) \right| \ln(1/q) \quad \forall \, k < 0.5 \quad (9)$$

*mbest* is mean of *pbest* of the swarm across all dimensions and $\Phi_{ij}$ is the local attractor of particle *i*. $\theta$, $q$ and $k$ are i.i.d. uniform random numbers distributed in [0,1]. $\beta$ is the contraction-expansion coefficient which is varied over the iterations as:

$$\beta = (1 - 0.1) \left( \frac{iteration_{max} - iteration_{current}}{iteration_{max}} \right) + 0.1 \quad (10)$$

Eq. (6) updates the *pbest* set and its minimum is set as *gbest*.

## IV. FUZZY C-MEANS QPSO (FCM QPSO)

In this approach, each particle is a *D* dimensional candidate solution in one of the *C* clusters that can be formally represented as the matrix *X*:

$$X = \begin{bmatrix} x_{11} & \cdots & x_{1D} \\ \vdots & \ddots & \vdots \\ x_{C1} & \cdots & x_{CD} \end{bmatrix} \quad (11)$$

A population of particles is randomly initialized and personal as well as global best positions are determined. Subsequently membership values are computed and a cost is assigned to each particle. The QPSO algorithm minimizes the cost associated with the particles through recursively calculating the mean best position using eq. (7), the membership values and cost function through eqs. (1) and (3) and updating the candidate cluster centre solution *X*. The algorithm is terminated if there is no improvement in the global best and the algorithm stagnates or if the preset number of iterations is exhausted. By using the stochastic and non-differentiable objective function handling capabilities of QPSO within the FCM algorithmic framework, the problem of stagnation in a local minima within a multidimensional search space is mitigated to an extent better than that possible with only the traditional FCM. The pseudocode of FCM QPSO is outlined below:

| Algorithm 1 FCM QPSO |
|---|
| 1: **for** each particle $x_i$ |
| 2:    *initialize position* |
| 3: **end for** |
| 4: *Evaluate membership values using eq. (1)* |
| 5: *Evaluate cost using eq. (3) and set pbest, gbest* |
| 6: **do** |
| 7:    *Compute mean best (mbest) position using eq. (7)* |
| 8:    **for** each particle $x_i$ |
| 9:      **for** each dimension $j$ |
| 10:        *Calculate local attractor $\Phi_{ij}$ using eq. (8)* |
| 11:        **if** $k \geq 0.5$ |
| 12:          *Update $x_{ij}$ using eq. (9) with '+'* |
| 13:        **else** *Update $x_{ij}$ using eq. (9) with '-'* |
| 14:        **end if** |
| 15:      **end for** |
| 16:    *Evaluate cost using eq. (3) and set pbest, gbest* |
| 17:    **end for** |
| 18: **while** *max iter or convergence criterion not met* |

## V. EXPERIMENTAL SETUP

### A. Parameter Settings

Learning parameters *C1* and *C2* are chosen as 2.05 and inertia weight *ω* in PSO is decreased linearly from 0.9 to 0.1 over the course of iterations to facilitate global exploration in the early stages and exploitation in the latter stages. The contraction-expansion parameter *β* is varied according to eq. (10) for QPSO.

### B. Datasets

Five well-known real datasets from the UCI Machine Learning Repository were used in analysis. These are:

1) Fisher's Iris Dataset consisting of three species of the Iris flower (Setosa, Versicolour and Virginica) with a total of 150 instances with 4 attributes each.

2) Breast Cancer Wisconsin (Original) Dataset consisting of a total of 699 instances with 10 attributes and can be classified into 2 clusters: benign and malignant.

3) Seeds Dataset consisting of 210 instances with 3 different varieties of wheat (Kama, Rosa and Canadian), each with 70 instances and 7 attributes.

4) Mammographic Mass Dataset consisting of 961 instances with 6 attributes and classified into two clusters: benign and malignant based on BI-RADS attributes and patient's age.

5) Sonar Dataset with 208 instances with 60 attributes and can be classified into either of 2 objects: mines or rocks.

### C. Performance Indices

The performance indices which provide insight into the clustering effectiveness are outlined below:

(a) Intercluster Distance: The sum of distances between the cluster centroids, larger values of which are desirable and imply a greater degree of non-overlapping cluster formation.

(b) Intracluster Distance: The sum of distances between data points and their respective parent cluster centroids, smaller values of which are desirable and indicate greater compactness of clustering.

(c) Quantization Error: The sum of distances between data points in a particular cluster and that parent cluster centroid, divided by the total data points belonging to that cluster, subsequently summed over all clusters and averaged by the number of data clusters.

Indices such as F-Measure and Accuracy for the datasets under test are calculated. The clustering algorithms are implemented in MATLAB R2016a with an Intel(R) Core(TM) i7-5500U CPU @2.40GHz. Experimental results for 10 trials are tabulated and are thereafter analyzed. Table 1 lists the datasets used in this paper.

## VI. RESULTS AND ANALYSIS

Tables 2 through 6 contain results on clustering performance reporting mean and standard deviations for the performance indices for QPSO FCM, QPSO K-Means and PSO K-Means and Figures 1 through 5 compare the accuracy of each algorithm over all datasets.

### Table 1. Data Set Information

|  | No. of Data Points | No. of Attributes | No. of Clusters |
|---|---|---|---|
| Iris | 150 | 4 | 3 |
| Breast Cancer | 699 | 10 | 2 |
| Seeds | 210 | 7 | 3 |
| Mammographic Mass | 961 | 6 | 2 |
| Sonar | 208 | 60 | 2 |

### Table 2. Comparison of Various Performance Indices for Iris Data Set

|  | Inter Cluster Distance | Intra Cluster Distance | Quantization Error | F Measure |
|---|---|---|---|---|
| QPSO FCM | 5.7312±0.0067 | 9.4608±0.0109 | 0.6414±0.0035 | 0.9133±0.0000 |
| QPSO K-Means | 6.1476±0.0330 | 8.9548±0.0075 | 0.6123±0.0360 | 0.9030±0.0021 |
| PSO K-Means | 6.1411±0.0934 | 9.0478±0.0927 | 0.6418±0.0176 | 0.8937±0.0063 |

### Table 3. Comparison of Various Performance Indices for Breast Cancer Data Set

|  | Inter Cluster Distance | Intra Cluster Distance | Quantization Error | F Measure |
|---|---|---|---|---|
| QPSO FCM | 13.3462±1.2050 | 146.0102±5.2887 | 3.8394±0.0410 | 0.9641±0.0024 |
| QPSO K-Means | 14.2993±0.4993 | 142.9908±1.3635 | 5.3048±0.0514 | 0.9627±0.0028 |
| PSO K-Means | 14.1413±0.7667 | 141.7168±1.2193 | 5.2737±0.0415 | 0.9616±0.0038 |

### Table 4. Comparison of Various Performance Indices for Seeds Data Set

|  | Inter Cluster Distance | Intra Cluster Distance | Quantization Error | F Measure |
|---|---|---|---|---|
| QPSO FCM | 10.0939±0.3889 | 25.5266±0.5904 | 0.6677±0.0063 | 0.8953±0.0124 |
| QPSO K-Means | 9.8110±0.0051 | 24.3534±0.0059 | 1.4835±0.0162 | 0.8995±0.0000 |
| PSO K-Means | 9.8460±0.2415 | 24.5255±0.2069 | 1.4949±0.0020 | 0.8979±0.0044 |

### Table 5. Comparison of Various Performance Indices for Mammographic Mass Data Set

|  | Inter Cluster Distance | Intra Cluster Distance | Quantization Error | F Measure |
|---|---|---|---|---|
| QPSO FCM | 21.6798±5.8964 | 275.2412±17.1193 | 6.2366±0.0386 | 0.6910±0.0070 |
| QPSO K-Means | 21.5716±0.0174 | 261.0328±0.0361 | 7.3978± 0.0000 | 0.6855±0.0000 |
| PSO K-Means | 21.3634±0.4812 | 261.8542±1.1712 | 7.4319± 0.0153 | 0.6851±0.0011 |

### Table 6. Comparison of Various Performance Indices for Sonar Data Set

|  | Inter Cluster Distance | Intra Cluster Distance | Quantization Error | F Measure |
|---|---|---|---|---|
| QPSO FCM | 0.7381±0.0916 | 19.2274±0.6306 | 0.6851±0.0077 | 0.5989±0.0345 |
| QPSO K-Means | 1.3264±0.0446 | 17.0267±0.1275 | 0.9329±0.2072 | 0.5702±0.0025 |
| PSO K-Means | 1.2750±0.0081 | 16.8552±0.0181 | 1.0002±0.1073 | 0.5421±0.0145 |

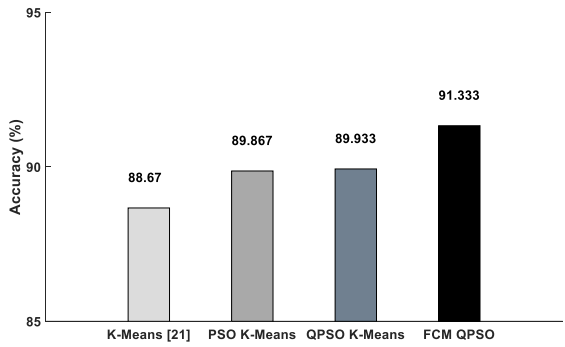

Figure 1. Accuracy of Algorithms on Iris

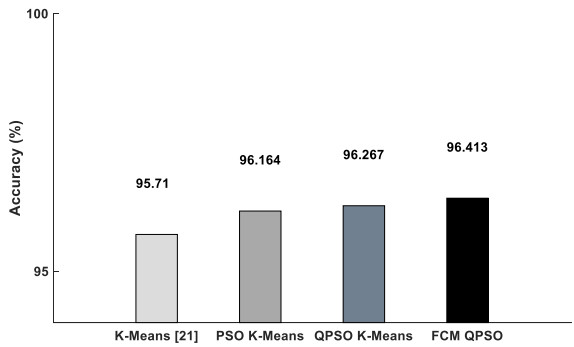

Figure 2. Accuracy of Algorithms on Breast Cancer

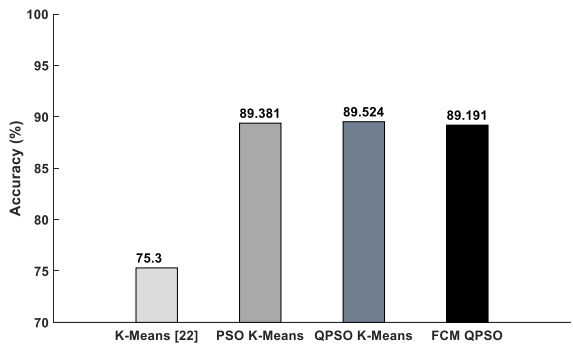

Figure 3. Accuracy of Algorithms on Seed

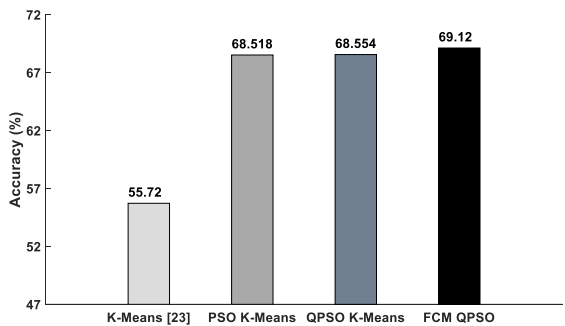

Figure 4. Accuracy of Algorithms on Mammographic Mass

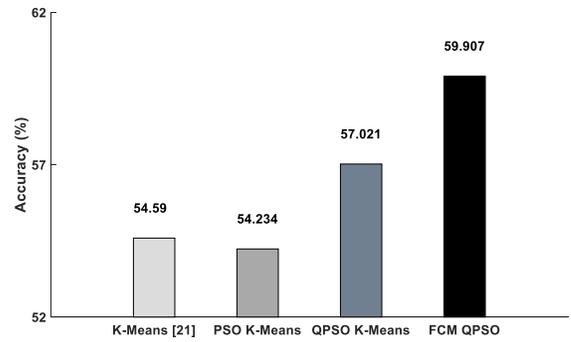

Figure 5. Accuracy of Algorithms on Sonar

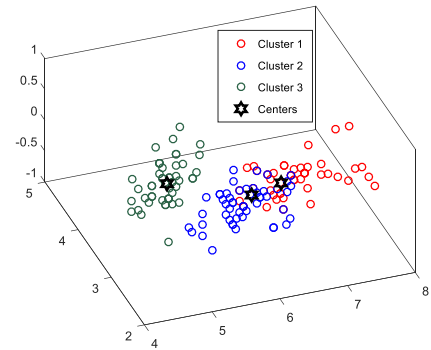

Figure 6. Clustering using FCM QPSO on Iris Dataset

Performance indicators such as Intercluster Distance, Intracluster Distance and Quantization Error computed from the results obtained in Tables 2-6 and that in Figures 1-5 imply that FCM QPSO has promising performance. Accuracy improvements of 1.556%, 0.151%, 0.825% and 5.061% respectively over QPSO K-Means are obtained on the Iris, Breast Cancer (Original), Mammographic Mass and Sonar data using FCM QPSO. On the Seed data, the accuracy drops by 0.371% and 0.212% for FCM QPSO when compared to QPSO K-Means and PSO K-Means, while recording an improvement of 18.447% over traditional K-Means.

The improvements in clustering accuracy and F-Measure obtained in case of FCM QPSO are at the expense of increased time complexity with respect to traditional K-Means based implementations. For instance, a typical FCM QPSO implementation with cluster numbers fixed a priori with the fuzzifier $m$ set as 2 results in approximately six times the computational cost as compared to QPSO K-Means when run on the Sonar dataset. Figure 6 shows a three dimensional partially representative classification of Iris Dataset into three distinct clusters along with the optimized cluster centres computed using FCM QPSO.

## VII. CONCLUSIONS AND FUTURE SCOPE

This paper makes an effort to compare and contrast the accuracy of hard and soft clustering techniques such as K-Means and Fuzzy C-Means upon hybridization with the standard, fully-connected quantum-behaved versions of the swarm intelligence paradigm of PSO on a number of datasets. FCM QPSO utilizes fuzzy membership rules of FCM and the guaranteed convergence ability of QPSO, thus avoiding stagnation in local optima in the multidimensional fitness landscape. Future work will analyze supervised approaches to mitigate the initial solution quality sensitivity in high dimensional datasets and aim at developing automatic techniques for detection of optimal cluster numbers and cluster centres in search spaces with reduced dimensionality.